\documentclass[a4paper, conference]{IEEEtran}
\usepackage{cite}
\usepackage{amsmath,amssymb,amsfonts}
\usepackage{graphicx}
\usepackage{textcomp}
\usepackage{xcolor}
\usepackage{booktabs}
\usepackage{textcomp}
\usepackage{dsfont}
\usepackage{color}
\usepackage[font={small}]{caption}
\usepackage{subcaption}
\usepackage{xfrac}
\usepackage{hyperref}
\usepackage[ruled, lined, linesnumbered, commentsnumbered, longend]{algorithm2e}
\newcommand{\pluseq}{\mathrel{+}=}

\DeclareMathOperator*{\argmin}{argmin}

\usepackage{booktabs}

\def\BibTeX{{\rm B\kern-.05em{\sc i\kern-.025em b}\kern-.08em
    T\kern-.1667em\lower.7ex\hbox{E}\kern-.125emX}}
\begin{document}

\title{Holistic Grid Fusion Based Stop Line Estimation}

\author{\IEEEauthorblockN{Runsheng Xu, Faezeh Tafazzoli, Li Zhang, Timo Rehfeld, Gunther Krehl, Arunava Seal}
\IEEEauthorblockA{Mercedes-Benz R\&D NA\\
Email: \{runsheng.xu, faezeh.tafazzoli, li.lz.zhang, timo.rehfeld, gunther.krehl, arunava.seal\}@daimler.com}}

\maketitle

\begin{abstract}
Intersection scenarios provide the most complex traffic situations in Autonomous Driving and Driving Assistance Systems. Knowing where to stop in advance in an intersection is an essential parameter in controlling the longitudinal velocity of the vehicle. Most of the existing methods in literature solely use cameras to detect stop lines, which is typically not sufficient in terms of detection range. To address this issue, we propose a method that takes advantage of fused multi-sensory data including stereo camera and lidar as input and utilizes a carefully designed convolutional neural network architecture to detect stop lines. Our experiments show that the proposed approach can improve detection range compared to camera data alone, works under heavy occlusion without observing the ground markings explicitly, is able to predict stop lines for all lanes and allows detection at a distance up to 50 meters.
\end{abstract}


\section{Introduction}
Navigating urban intersections is one of the most challenging tasks for self-driving cars (SDC). Because they are both confusing and potentially dangerous due to complicated topologies and traffic regulations \cite{Marta_stopline, intersection_protocol}, it is of fundamental importance to find solutions that can increase the safety of the traffic participants, to prevent and reduce the number of accidents and fatalities in intersections \cite{intersection_prove}.

In general, the vehicle\textquotesingle s situational awareness is directly related to the lookahead distance and lookahead time of the lane and topology estimates to localize ego-vehicle position and thereby determine a safe travel speed. Estimating all parameters of an intersection, e.g. stop lines, crosswalks, yield relations and possible paths through the intersection based on sensor data alone is very difficult; in particular when the SDC is still approaching the intersection and thus sensor data is very sparse. Most current approaches to self-driving solve this problem by localizing the vehicle relative to a high-definition (HD) map that contains all the static information about the environment in high detail. In this way, the SDC precisely knows where to stop when approaching the intersection and can plan a smooth trajectory accordingly. However, HD maps come with the potential risk of not being up-to-date, so a sensor-driven reasoning to complement and validate HD map is highly desirable.

\begin{figure}[tbp]
\centering
    \begin{subfigure}[b]{0.32\linewidth}
         \centering
         \includegraphics[width=\linewidth]{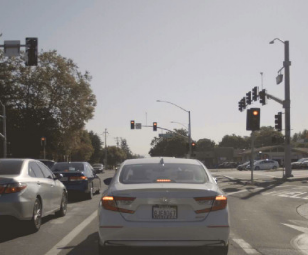}
         \caption{}
         \label{fig:corner_cases_occlusion}
     \end{subfigure}
     \begin{subfigure}[b]{0.32\linewidth}
         \centering
         \includegraphics[width=\linewidth]{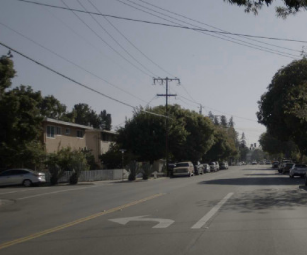}
         \caption{}
         \label{fig:corner_cases_crosstraffic}
     \end{subfigure}
     \begin{subfigure}[b]{0.32\linewidth}
         \centering
         \includegraphics[width=\linewidth]{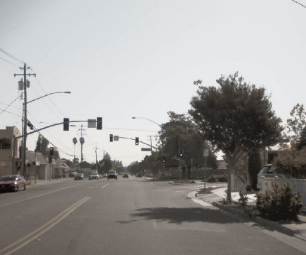}
         \caption{}
         \label{fig:corner_cases_farfield}
     \end{subfigure}
\caption{Challenging urban scenarios for stop line detection: (a) Occlusion on stop line. (b) The stop line for cross-traffic on the left can easily be confused with a regular lane marking. (c) The stop line can barely be seen in the camera from a distance of around 30 meters.}
\label{fig:corner_cases}
\end{figure}

\begin{figure}[htbp]
\centering
    \begin{subfigure}[b]{\linewidth}
         \centering
         \includegraphics[width=\linewidth]{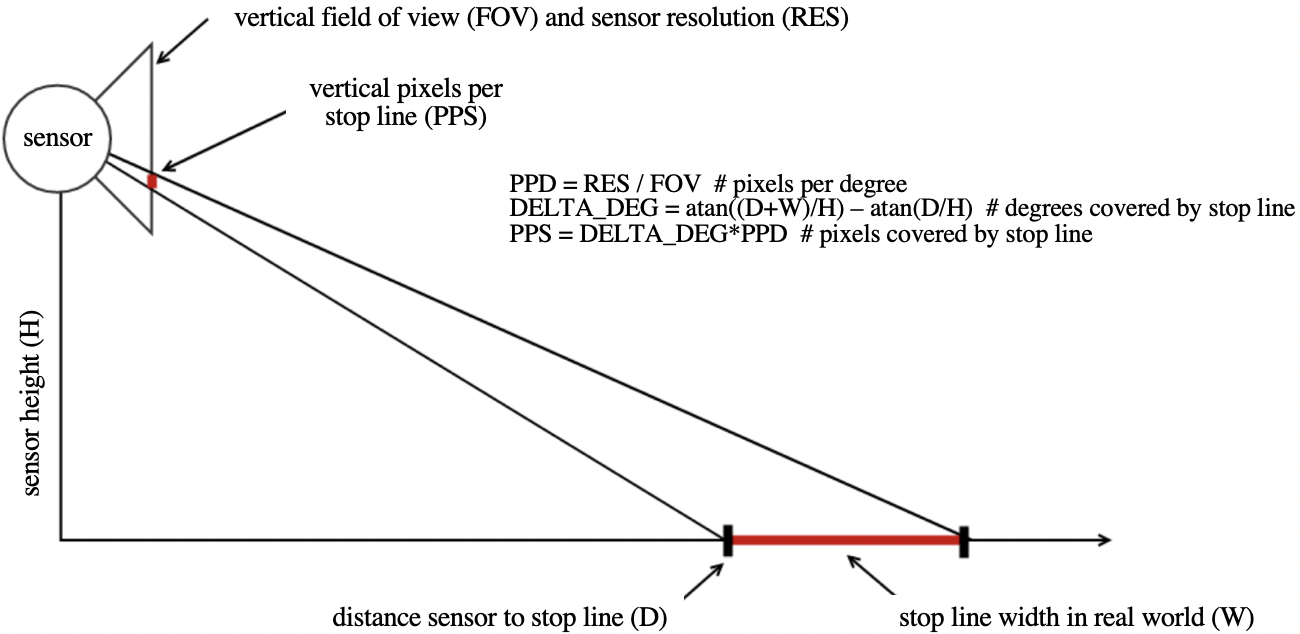}
         \caption{Simplified geometry model}
         \label{fig:sensor_model}
     \end{subfigure}
     \begin{subfigure}[b]{\linewidth}
         \centering
         \includegraphics[width=\linewidth]{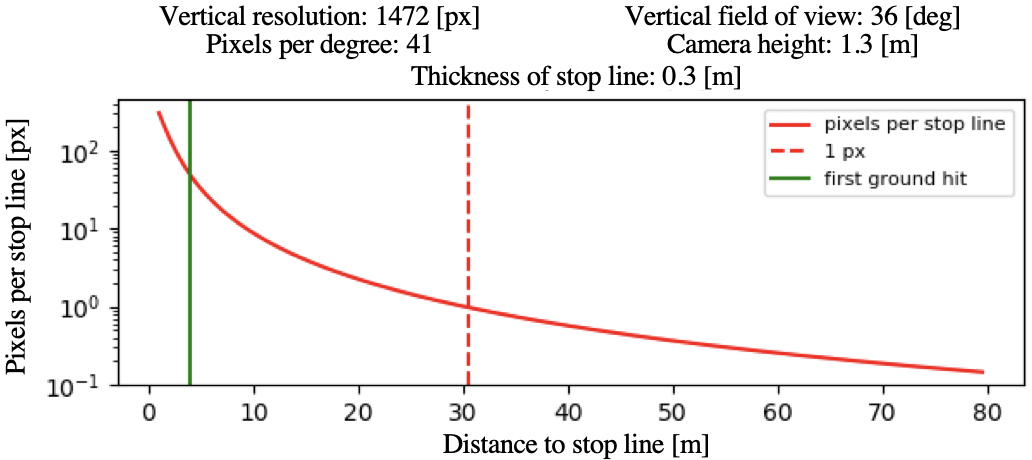}
         \caption{Relationship between distance and pixels number in camera sensor}
         \label{fig:distance_pixel}
     \end{subfigure}
\caption{(a) Model to compute the number of pixels visible in a sensor over distance. (b) Theoretical size of stop lines in pixels (vertical), over distance from ego to stop line. The camera configuration of the vehicle used for our data collection is also demonstrated.}
\label{fig:sensor_pixels}
\end{figure}

Most approaches to stop line detection rely only on camera images \cite{damage_stopline} and explicitly try to detect the stop line that is painted onto the road. A few difficult scenarios for camera-based stop line detection are depicted in Figure \ref{fig:corner_cases}.
Unfortunately, even with high resolution cameras and telephoto lenses it is very difficult to detect stop lines further than 30 meters ahead. This is because of the shallow angle between the camera\textquotesingle s line of sight and the road surface, as shown in Figure \ref{fig:sensor_model}. For a typical camera resolution and field-of-view the stop line is only around 1 pixel high in the image at around 30 meters, c.f. Figure \ref{fig:sensor_pixels}. Even with perfect marking conditions and high contrast to the neighboring asphalt, robustly detecting a 1 pixel signal is very challenging. At the same time, a comfort deceleration in urban traffic is around $3\; m/s^2$ \cite{deceleration} and thus a vehicle traveling at $35\;miles/h$ would require a stable stop line detection at least $53$ meters ahead\footnote{Assuming a conservative overall system latency (from perception to action) of $t = 0.8$ seconds, and using a simplified model for required detection distance of $d = \frac{v^2}{2a} + vt$ (not considering jerk limits).}.
High-resolution lidar sensors are able to pick up road markings at slightly higher distances, so they can increase detection range slightly, but beyond that no sensor can pick up explicit marking signals anymore.

To tackle this problem, instead of trying to explicitly detect stop lines painted onto the ground, we propose a learning-based pipeline that employs a variety of input signals from multiple sensors fused into a grid map representation that contain explicit and implicit cues about the existence of stop lines (Figure \ref{fig:gridmap}). In that sense, our work can also be defined as generally detecting the presence of an intersection and then deriving the expected stop lines of that intersection. Hence, our approach is able to not only estimate stop lines for the ego-vehicle, but it can also infer stop lines for other lanes, i.e. oncoming and cross traffic. This would provide viable information to the motion planner to boost the prediction of other agents behavior in the scene. Note that our definition of stop line is very general and covers all the areas where the SDC and other agents should stop in an intersection including crosswalks. The contribution is manifold:
\begin{itemize}
\item We show that utilization of the information obtained by different sensors, which avoids the perceptual limitations and uncertainties of a single sensor, forms a more comprehensive perception of the environment, and increases the robustness and accuracy of stop line detection beyond observable markings on the ground.
\item We propose a joint learning method for segmentation of small objects such as stop lines without additional computation during inference. From the binary segmentation mask, we create a distance map and a direction map that embed continuous implicit information of the underlying spatial structure of the stop lines. These maps will be learnt with the segmentation mask jointly in the training and provide auxiliary loss to enhance the feature representation of the encoders. 
\item In order to increase the interpretability of neural network results for safety-critical applications like stop line detection, following \cite{pay_attetnion}, we present a visual analysis of what information attracts the learned model most.
\end{itemize} 
\section{Related Work}
Reliable and accurate detection of stop lines has been a long-standing problem in both self-driving cars and Advanced Driver Assistance Systems (ADAS) \cite{Marta_stopline,stop_line_important_tak,stop_line_important_kim}. Existing methods mainly use computer vision techniques to tackle the problem and fall into two main categories: traditional computer vision based methods and deep learning based methods.

In \cite{fast_sym}, Suhr et al. perform simple edge detection and RANSAC to get stop line candidates. In the next step, they extract HOG features around the candidate areas and apply a TER-based classifier \cite{ter} to distinguish stop lines. Both \cite{Marta_stopline} and \cite{aru_stopline} apply a Hough transformation to retrieve image features and then follow a learning-based and rule-based model respectively to classify stop lines. Similarly, \cite{Lee_curbe} uses Canny edge detection and thresholding to find horizontal stop lines. Li et al. \cite{IPM_stopline} proposed an Inverse Perspective Mapping (IPM) transformation on the camera image to extract an accurate region of interest. Subsequently, stop lines are extracted via horizontal morphological operations. To track stop lines robustly over time, \cite{Tech_vision} uses a spatial filter to find stop line candidates and then applies a Kalman Filter to track and stabilize the position estimate. To eliminate the prior assumptions about ideal shapes or positional relations, \cite{damage_stopline} developed a system for detecting stop lines on roads with damaged paint. Although their proposed method achieves good results, it still is constrained by the requirement of partial visibility of the ground markings, and hence is not robust to occlusion.

To tackle the limited expressiveness and lack of generalization of traditional computer vision feature extractors, recently some deep learning based methods have been proposed in this context. In \cite{stop_deep}, Lin et al. combine Cascade-Adaboost and Convolutional Neural Networks (CNNs) together as stop line detector. Their method offers a better false alarm rate than the previous traditional approaches. Similarly, in \cite{lin2018one} ResNet50 is used as backbone to detect stop line bounding boxes. They are able to achieve fair accuracy with good inference speed. A vanishing point guided network is proposed in \cite{VPGNet} which predicts lanes and road markings such as stop lines. This approach is claimed to be robust under different illumination and weather conditions.

Although all of the above methods have made different breakthroughs in the context of stop line detection, they are all based on only camera sensor data. This limits their detection range, resulting in lower robustness in complex urban environments in addition to poor functioning under partial or full occlusion.
\section{Methodology}
\begin{figure}[htbp]
\centering
\includegraphics[width=\linewidth]{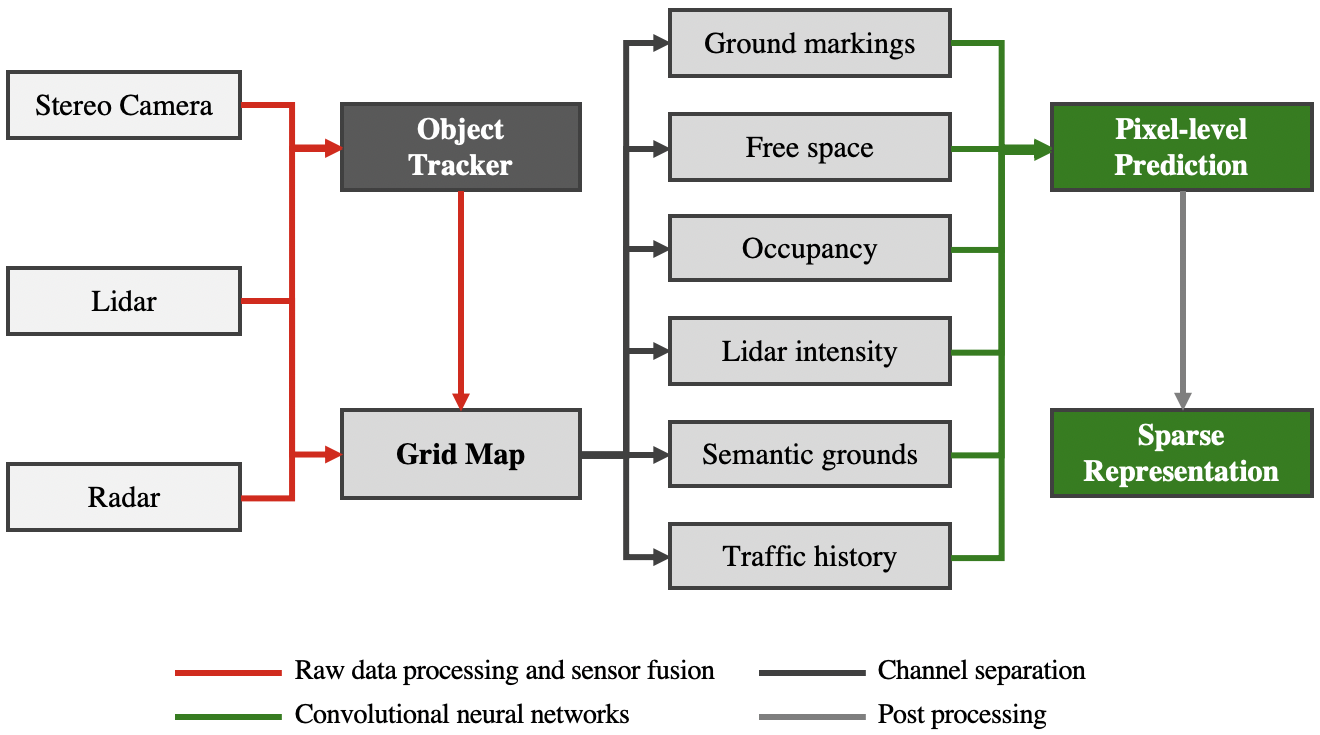}
\caption{Overview of the proposed methodology.}
\label{fig:overview}
\end{figure}

To enhance the stop region detection in order to ensure the safety and better path planning of the autonomous vehicle, we herein propose a learning-based pipeline which uses a multi-sensory scheme demonstrated in Figure \ref{fig:overview}. The SDC is configured with stereo camera and lidar to collect different sensor raw measurements. These measurements are processed individually as the first step to obtain both static environment information and dynamic object information. The system is also equipped with radar to enable tracking of the fused information obtained from dynamic objects detected through each sensor modality and acquire accumulated trajectories of the corresponding objects over time. The resulting trajectories along with the static environment information are formulated as a grid map representation. Next, we apply an encoder-decoder neural network on the grid map to generate a segmentation mask. Finally, the mask will be post-processed to extract the sparse representation of stop lines to assist behavior planning.
\subsection{Input Parameterization}
\label{input_parameterization}
In the automotive context, an Occupancy Grid Map (OGM) \cite{elfes1990occupancy} presents a low-level abstraction of the driving environment in form of a collection of cells, usually of the order of centimeters, paired with their occupancy probabilities. The location of empty spaces and obstacles is estimated from the spatial disposition and the occupancy probabilities of cells. This enables the fusion of heterogeneous sensors with different characteristics mounted at different locations of the vehicle. The fusion scheme makes use of the Bayesian theory to build such probabilistic environment model by taking into account the sensor uncertainties in addition to Dempster-Shafer theory \cite{shafer1976mathematical} to handle conflicting information. This makes occupancy grids suitable for representing unstructured environments SDC faces with in urban driving.

\begin{figure*}[htbp]
\centering
    \begin{subfigure}[b]{0.24\linewidth}
         \centering
         \includegraphics[width=0.88\linewidth]{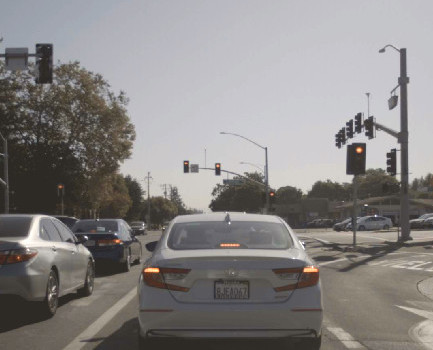}
         \caption{Camera snapshot}
         \label{fig:corner_cases_occlusion}
     \end{subfigure}
     \begin{subfigure}[b]{0.24\linewidth}
         \centering
         \includegraphics[width=0.88\linewidth]{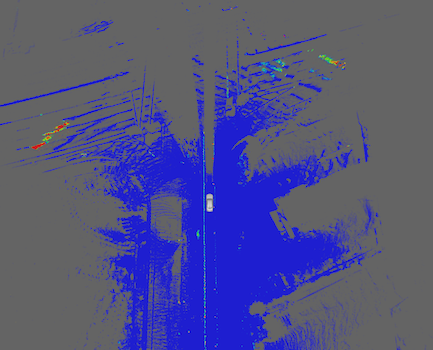}
         \caption{Lidar intensity}
         \label{fig:corner_cases_crosstraffic}
     \end{subfigure}
     \begin{subfigure}[b]{0.24\linewidth}
         \centering
         \includegraphics[width=0.88\linewidth]{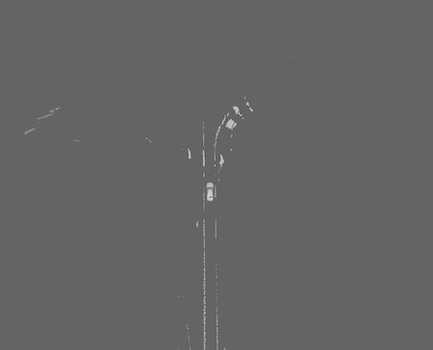}
         \caption{Ground markings}
         \label{fig:corner_cases_farfield}
     \end{subfigure}
      \begin{subfigure}[b]{0.24\linewidth}
         \centering
         \includegraphics[width=0.88\linewidth]{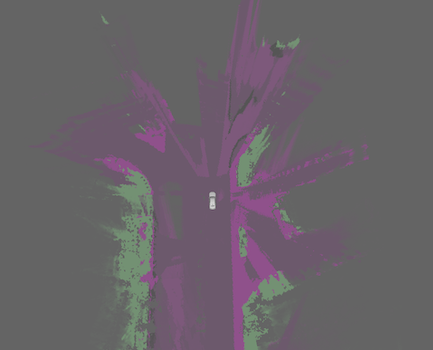}
         \caption{Semantics ground}
         \label{fig:corner_cases_farfield}
     \end{subfigure}
     \begin{subfigure}[b]{0.24\linewidth}
         \centering
         \includegraphics[width=0.88\linewidth]{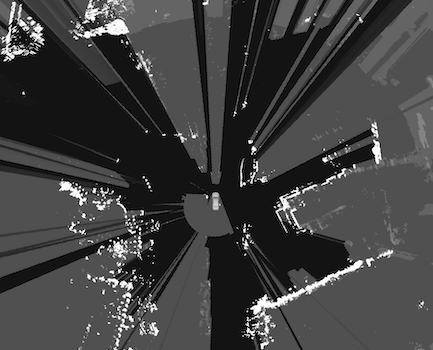}
         \caption{Occupancy}
         \label{fig:corner_cases_farfield}
     \end{subfigure}
     \begin{subfigure}[b]{0.24\linewidth}
         \centering
         \includegraphics[width=0.88\linewidth]{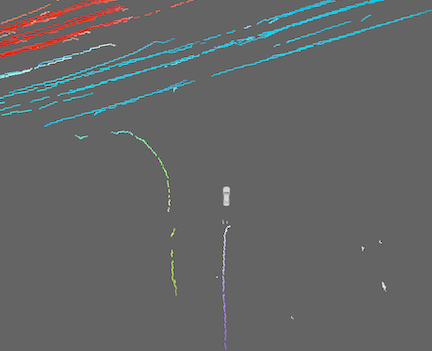}
         \caption{Traffic history}
         \label{fig:corner_cases_farfield}
     \end{subfigure}
      \begin{subfigure}[b]{0.24\linewidth}
         \centering
         \includegraphics[width=0.88\linewidth]{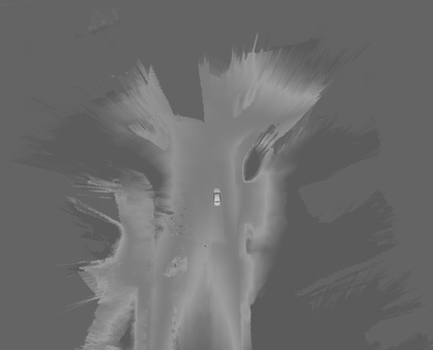}
         \caption{Elevation map}
         \label{fig:corner_cases_farfield}
     \end{subfigure}
     \begin{subfigure}[b]{0.24\linewidth}
         \centering
         \includegraphics[width=0.88\linewidth]{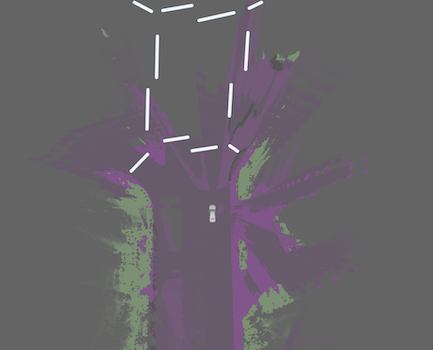}
         \caption{Stop line ground truth}
         \label{fig:corner_cases_farfield}
     \end{subfigure}
\caption{A concrete example indicating different channels of the grid map.}
\label{fig:gridmap}
\end{figure*}

Our system integrates the fused information as vehicle-centered grid map encoding both geometric and semantic information in multiple layers, including occupancy, ground markings, lidar intensity, semantics ground, and traffic history. Figure \ref{fig:gridmap} demonstrates an example of these layers. Each channel is dedicated to a semantically different source of information; lidar intensity depicts the projected 3D lidar cloud points; ground semantics illustrates the segmentation for road surface and grass, providing information on the lane positioning; traffic history defines presence of cross traffic by outlining the accumulated object trajectories whose color represent different velocity directions; elevation map delivers the information of elevation relative to ground for each pixel; and occupancy provides information about static infrastructure such as buildings that delimit the road topology. Such grid map which is essentially a bird\textquotesingle s eye view (BEV) of the ego-vehicle\textquotesingle s environment form the input to our pipeline, referred to as $I \in \mathbb{R}^{C \times H \times W}$, where $C$ is the number of aforementioned grid map layers.
\subsection{Feature Representation Enhancement via Joint Learning}
\label{feature_enhancement}
Stop line detection is first formulated as a semantic segmentation task. Given the input grid map representation tensor $I$, the target is to generate a segmentation map $S \in \mathbb{R}^{1\times H \times W} $ in which each pixel $(i,j)$ has a binary label to indicate the stop line existence.\\
From the grid map, we retrieve different information channels and pass them to a network withe an architecture similar to UNet \cite{UNet} to perform distance map regression and stop lines segmentation. 
\subsubsection{Network Architecture Overview}
\begin{figure*}[htp]
    \centering
    \includegraphics[width=0.9\linewidth]{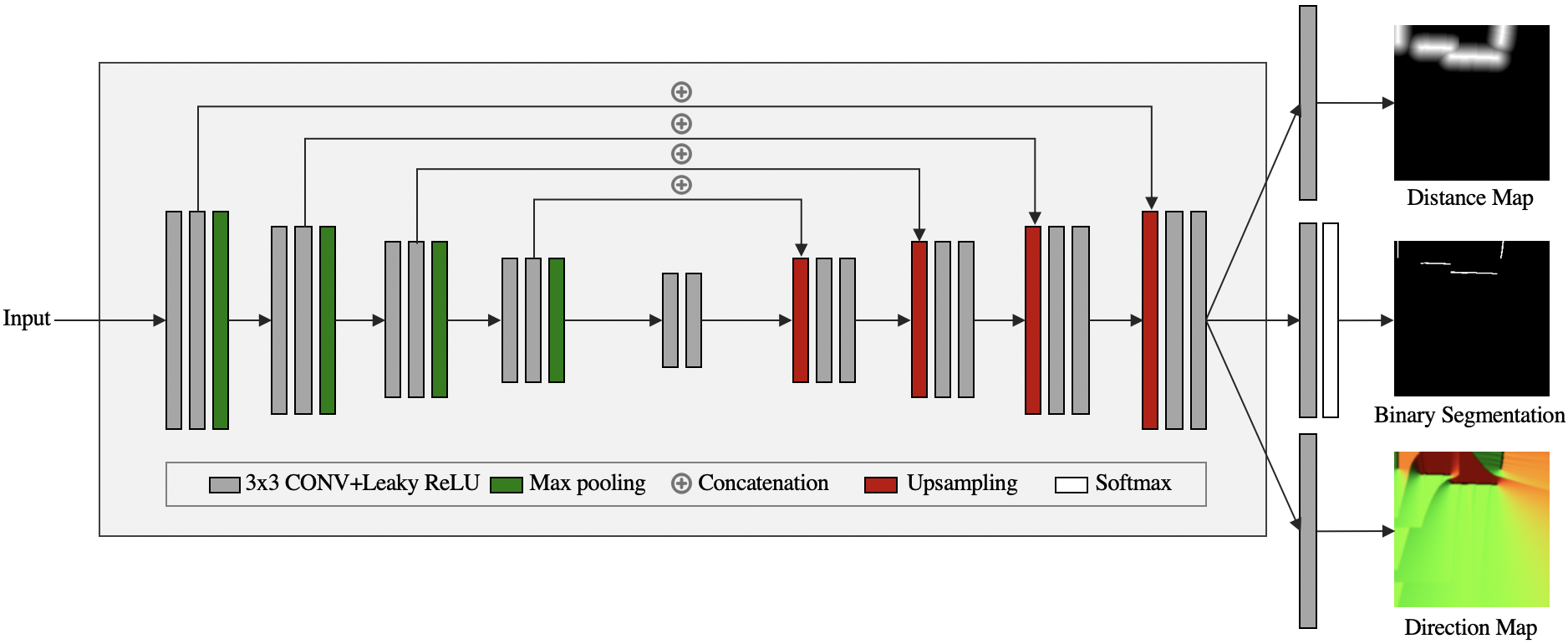}
\caption{Overview of the proposed joint learning pipeline and its outputs. The direction map here is shown as a flow field similar to \cite{flow_field}.}
\label{fig:network_overview}
\end{figure*}

 Segmentation binary masks usually provide limited information in the gradient back-propagation when the data is severely biased because the pixels of the masks can just indicate their association to a certain class. Thus, similar to \cite{distance_transform}, we use Signed Distance Transform (SDT) \cite{SDT} on the binary mask to retrieve equivalent continuous representations referred to as distance map, which can offer spatial proximity information. We, also, predict the direction map to provide spatial direction information and assist the neural network to better understand the underlying spatial structure of the data. As depicted in Figure \ref{fig:network_overview}, these two masks will add auxiliary regression loss through joint learning during training. To avoid additional computation cost, during inference only the segmentation mask will be generated.\\
 We have used UNet\cite{UNet} as the backbone for the joint learning because of its capability to retain both low level details and high level semantic information\cite{residual_net}. 
\subsubsection{Distance and Direction Map}
To provide richer spatial proximity information, SDT \cite{distance_transform} is used to estimate the distance map. Assume $X_{ij}$ is the 2D coordinate vector of pixel $(i,j)$ in input image, $M$ is the foreground mask representing the stop line areas, then the distance value $d_{ij}$ in distance map $D \in \mathbb{R}^{1\times H \times W}$ and the closest foreground point $F_{ij}$ to this pixel are respectively defined as:
\begin{align}
    \forall{i,j} \quad d_{ij} = \begin{cases}
                d_{thresh}, \quad if \; X_{ij} \subset M \\ 
                d_{thresh} - min_{z \subset M}(||X_{ij}-z||) \quad if \; X_{ij} \notin M\\
    \end{cases}
\end{align}
\begin{align}
        \forall{i,j} \quad F_{ij} = \begin{cases}
                X_{ij}, \quad if \; X_{ij} \subset M \\ 
                \argmin\limits_{z \subset M}(||X_{ij}-z||) \quad if \; X_{ij} \notin M\\
        \end{cases}
\end{align}
where $d_{thresh}$ is a positive integer.

We use a sequential algorithm \cite{a_linear} to find the closest foreground point for each pixel in linear time. The negative values in distance map will be clipped to 0 and positive ones will be normalized into $(0,1]$. Figure \ref{fig:network_overview} shows an example of the generated distance map.

Since the closest foreground point $F_{ij}$ is retrieved for each pixel $(i,j)$ in the input image, the offsets $x_d$ and $y_d$ on both axis can be calculated to generate a direction map $E \in \mathbb{R}^{2\times H \times W} $. This direction map will implicitly provide the continuous spatial direction information from each pixel to its closest stop line point. An example of the direction map is displayed in Figure \ref{fig:network_overview}.

We use cross-entropy loss for segmentation task and $L_2$ loss for regression task.
\begin{equation}
\begin{split}
L_{seg} = & L_{cross}(S_{pred}, S_{gt}) + \\
          & \lambda_1 L_2(D_{pred}, D_{gt}) + \lambda_2 L_2(E_{pred}, E_{gt})
\end{split}
\end{equation}
The hyper-parameters $\lambda_1$ and $\lambda_2$ are both set to $0.5$ during training.
\subsection{Sparse Representation}
In the next step, the segmentation map $S$ is processed to extract the sparse representation of the existing stop lines. For each stop line $l_i$ four properties $p_{start},\;p_{end},\;length,\;slope$ are estimated which represent the 2D image coordinates of the start and end points of the lines in the image space, the euclidean distance between the start and end point, and the slope of the stop line, respectively. For that, Connected-Component Labeling is first applied on segmentation map $S$ to divide the connected positive predictions into different groups. Since each group is a cluster of 2D points, we apply Principal Component Analysis (PCA) to obtain the principal axis of each cluster and thus gain the slope of the underlying stop line. The leftmost and rightmost coordinates of each group are projected to the slope axis to estimate the start and end points of the line segment as part of the stop line sparse representation.

However, object segmentation mask does not always yield continuous results. Two lines originating from the same stop line might be regarded as two different groups because of the segmentation truncation. Therefore, we perform a simple refining algorithm based on pairwise comparison of stop lines $l_i$ and $l_j$ to apply dilation and potentially removing the duplicate line if they meet the following conditions: $ dist(l_i, l_j) < d_{thresh}$ and $angle(l_i, l_j) < a_{thresh}$, having
\begin{align}
    dist(l_i,l_j) = \frac{ \sum_{k=1}^{n}{PerpDist}({p_{i_k}}, l_j)}{n}
\end{align}
\begin{align}{
    angle(l_i,l_j) = \arctan{|\frac{slope_{i} - slope_{j}}{1+slope_{i}slope_{j}}|}}
\end{align}
where ${p_{i_1}}, {p_{i_2}}, ...., {p_{i_n}}$ denotes the $n$ interpolation points between the start point and end point of stop line $l_i$, and ${PerpDist}({p_{i_k}}, l_j)$ represents the perpendicular distance between the point $p_{i_k}$ and line $l_j$. The parameters $d_{thresh}$ and $a_{thresh}$ are experimentally chosen to $0.3\;m$ and $8\;degree$ in our experiments.

After retrieving the sparse representation of both prediction $L_p$ and ground truth $L_g$ stop lines, to perform evaluation and estimate the accuracy of model we use Algorithm \ref{algorithm:match} to match ground truth data, provided through HD map, with predicted stop lines. As criterion for a match we do the association with the closest ground truth stop line and require both lines to overlap and align, relaxing the constraints with a margin of error $a_{thresh}$ for orientation. 
\begin{algorithm}
    \SetAlgoLined
    \SetKwInOut{KwIn}{Input}
    \SetKwInOut{KwOut}{Output}
    \KwIn{List of stop line predictions $L_p$ and ground truth $L_g$, $a_{thresh}$.}
    \KwOut{Number of matched detections $n_{pos}$, number of false alarms $n_{neg}$, and total distance error $e_{total}$.}
    $n_{gt}=length(L_g),\, n_{pred}=length(L_p)$\\
    $n_{pos}=0,\, n_{neg}=0,\, e_{total}=0$
    
    \For{$i \leftarrow 1$ \KwTo $n_{gt}$}{
        $d_{min}\gets inf,\, idx\gets -1$
        
        \For{$j \leftarrow 1$ \KwTo $n_{pred}$}{
            \If{$l_{g_{i}}$ overlaps with $l_{p_{j}}$ and $dist({l_{g_{i}}},l_{p_{j}}) < d_{min}$ and $angle({l_{g_{i}}},l_{p_{j}}) < a_{thresh}$}{
                $d_{min}$ $\gets$ {$dist({l_{g_{i}}}, l_{p_{j}})$}, $idx$ $\gets$ $j$
            }
        }
        \If{$idx \neq -1$}{
            $n_{pos}$ $\pluseq$ $1$, $e_{total} \pluseq d_{min}$
            
            $del$ $l_{p_{idx}},\; l_{g_{i}}$
        }
    }
    $n_{neg}\gets length(L_p)$
    
    \KwRet{$n_{pos}$, $n_{neg}$, $e_{total}$}
    \caption{Stop line Association}
    \label{algorithm:match}
\end{algorithm}
\section{Experiments}
\subsection{Dataset}
The experiments are done on datasets recorded from Santa Clara county, United State. We use the HD map data as ground truth to train our model in a self-supervised way. The data was collected from multiple passes of the autonomous vehicle through out 10 days, covering various traffic scenarios. The training set consists of $195$ minutes driving duration and $65$ miles of driving stretch. To evaluate the generalization capability of our model, a recording from downtown San Jose, CA is used as test set. This dataset contains a $20$ minute drive along $7$ miles of driving distance. Overall, the curated data for this paper contains around 9k frames for training, 1k frames for validation and 1k frames for testing.
\subsection{Implementation Details}
The model is trained with learning rate of $1e^{-4}$ and $L2$ weight decay of $2e^{-4}$ along with batch size of $16$. Adam optimizer with the same custom class weighing scheme as \cite{enet} for binary cross entropy loss is applied. We use image rotation to augment data 3 times and train the network for $250k$ iterations. 

The end-to-end inference time of the system is $27ms$ on average on a GeForce GTX 980 Ti GPU, which is on par with real-time performance.
\subsection{Experimental Setup}
To have a comprehensive evaluation of our approach, based on the distance of the stop lines from the ego-vehicle, we divide the dataset into 5 different categories in the following ranges: 0-10, 10-20, 20-30, 30-40, and 40-50 meters.\\
As evaluation metrics we use Precision, Recall and F1 score for general detection performance and mean absolute error (MAE) for precision positioning error. We use a grid map encoding with the resolution of $26cm\times 26cm$ per cell. Therefore, we use meter to measure the position MAE instead of pixels to make the results more intuitive. 
\subsection{Quantitative and Qualitative Results}
\begin{figure}[htbp]
\centering
     \begin{subfigure}[b]{0.32\linewidth}
         \centering
         \includegraphics[width=\linewidth]{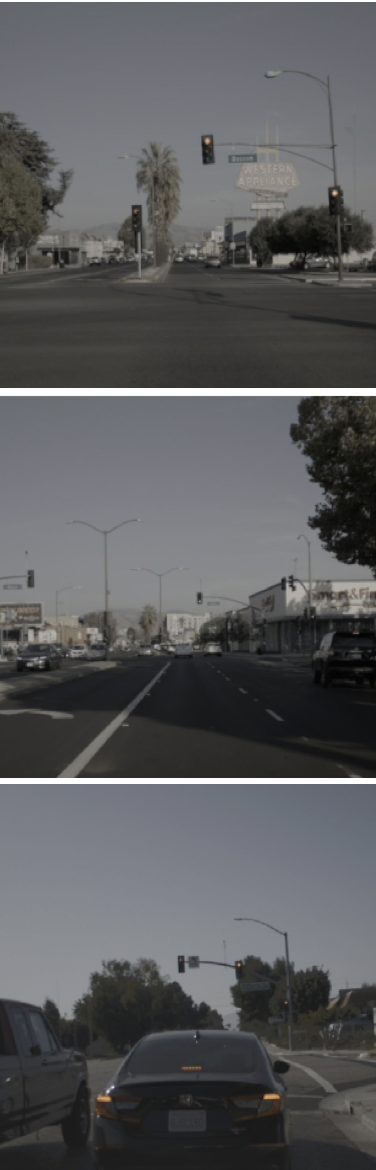}
         \caption{Stereo camera}
         \label{fig:qualitative_results_1}
     \end{subfigure}
           \begin{subfigure}[b]{0.32\linewidth}
         \centering
         \includegraphics[width=\linewidth]{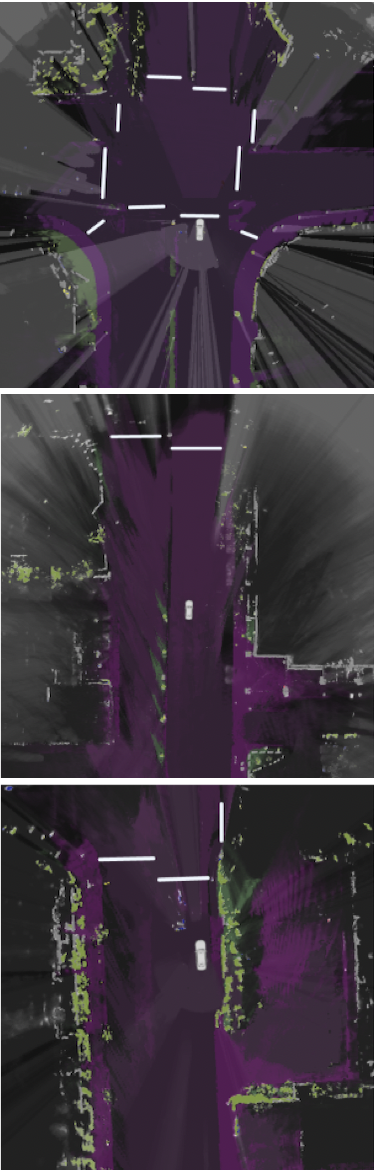}
         \caption{Ground truth}
         \label{fig:qualitative_results_2}
     \end{subfigure}
      \begin{subfigure}[b]{0.32\linewidth}
         \centering
         \includegraphics[width=\linewidth]{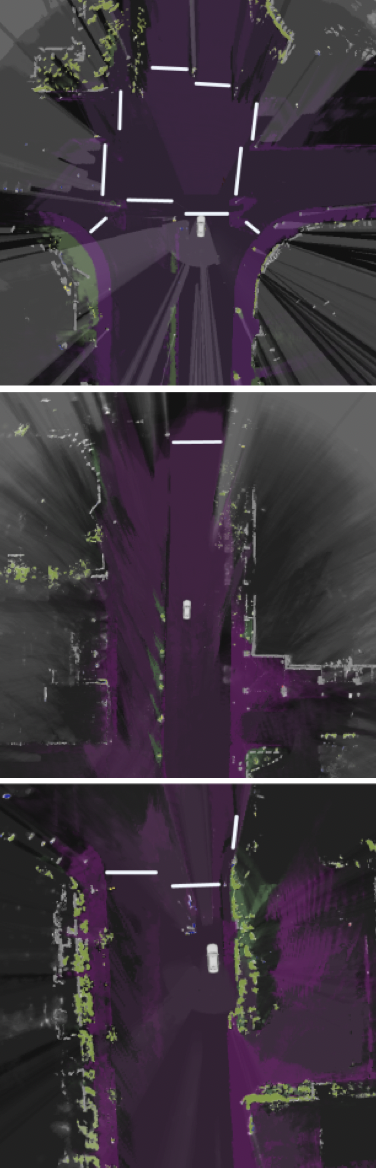}
         \caption{Predictions}
         \label{fig:qualitative_results_3}
     \end{subfigure}
\caption{Qualitative results of our stop line estimation approach. The front camera image is shown as reference (left). The results indicate that our approach is able to detect stop lines for all lanes (top), at high distances (middle), and during occlusion (bottom).}
\label{fig:qualitive_result_over}
\end{figure}

Figure \ref{fig:qualitive_result_over} displays examples of inferred stop lines overlayed on combined input channels. The results depict different scenarios with respect to intersection topology and distance of ego-vehicle from the stop line area. Table \ref{table:method_ablation} demonstrates the importance of joint learning in achieving a model capable of detecting all stop lines present in the scene regardless of road structure. Even though the test set has a very different road structure compared to the training set, our model can still achieve a decent performance in close range under severe occlusions and is able to detect stop lines 50 meters ahead with a F1 score more than $60\%$ and average error less than $60$ centimeters.
\subsection{Ablation Study}
\subsubsection{Distance and Direction Map Analysis}
As discussed in \ref{feature_enhancement}, we augmented stop line segmentation mask with auxiliary losses to also predict the distance map and direction map. Table \ref{table:method_ablation} shows that the proposed joint learning approach effectively improves both F1 and MAE score across all ranges. Our method is able to achieve $86.1\%$ F1 score with $0.26m$ MAE in close range and $60.3\%$ F1 score with $0.54m$ MAE at around $50m$ distance.
\begin{table*}[htbp]
\centering
\resizebox{\linewidth}{!}
{\begin{tabular}{lcccccccccccccccccccccccc}
\toprule
Distance & \multicolumn{4}{c}{\textbf{10m}} & & \multicolumn{4}{c}{\textbf{20m}} & & \multicolumn{4}{c}{\textbf{30m}} & & \multicolumn{4}{c}{\textbf{40m}} & & \multicolumn{4}{c}{\textbf{50m}}\\
\cmidrule{2-5} \cmidrule{7-10} \cmidrule{12-15} \cmidrule{17-20} \cmidrule{22-25}
         & Pr. & Rec. & F1 & MAE & & Pr. & Rec. & F1 & MAE & & Pr. & Rec. & F1 & MAE & & Pr. & Rec. & F1 & MAE & & Pr. & Rec. & F1 & MAE\\
\cmidrule[1.5pt]{2-5} \cmidrule[1.5pt]{7-10} \cmidrule[1.5pt]{12-15} \cmidrule[1.5pt]{17-20} \cmidrule[1.5pt]{22-25}
S        & 85.5 & 74.9 & 80.0 & 0.32 & & 83.7 & 67.3 & 74.7 & 0.48 & & 82.7 & 63.8 & 72.0 & 0.48 & & 74.5 & 57.8& 65.2 & 0.54 & & 67.3 & 51.4 & 58.3 & 0.72\\
S+E      & 93.4 & 75.9 & 83.8 & 0.29 & & 80.9 & 71.5 & 76.0 & 0.34 & & 81.0 & 70.7 & 75.5 & 0.56 & & 71.8 & 63.7 & 67.4 & 0.57 & & 67.5 & 51.6 & 58.6 & 0.66\\
S+D      & 93.7 & 77.4 & 84.8 & 0.36 & & 81.3 & 73.5 & 77.2 & 0.30 & & 82.0 & 71.3 & \textbf{76.3} & 0.48 & & 67.7 & 58.8 & 63.0 & 0.52 & & 62.7 & 45.9 & 53.8 & 0.62\\
S+E+D    & 94.1 & 79.4 & \textbf{86.1} & \textbf{0.26} & & 83.8 & 77.4 & \textbf{80.6} & \textbf{0.28} & & 80.1 & 72.1 & 76.0 & \textbf{0.47} & & 74.7 & 66.0 & \textbf{70.1} & \textbf{0.48} & & 62.8 & 56.2 & \textbf{60.3} & \textbf{0.54}\\
\bottomrule
\end{tabular}}
\caption{Quantitative analysis of the importance of joint learning, on the inclusion of Segmentation mask (S), Distance map (D), and Direction map (E).}
\label{table:method_ablation}
\end{table*}
\subsubsection{Input Channels Analysis}
In order to validate the importance of data fusion and implicit features defined in \ref{input_parameterization}, we first conduct an experiment that only takes single sensor data as input and thereby focuses on the corresponding sensor readings of the ground that provide explicit road marking information. Ground markings and lidar intensity are generated individually for this purpose by stereo camera and lidar respectively. Hence as a comparison with other existing approaches relying only on one sensor, we perform an ablation study on those two channels. Table \ref{table:singlesensor} shows that compared to the results provided in Table \ref{table:method_ablation}, for both sensors the performance decreases in all distance ranges. That is even more remarkable in case of far distance stop lines where it degrades dramatically around $40\%$ percentage for camera in the range of 40-50 meters.
\begin{table*}[htbp]
\centering
\resizebox{\linewidth}{!}
{\begin{tabular}{lcccccccccccccccccccccccc}
\toprule
Distance & \multicolumn{4}{c}{\textbf{10m}} & & \multicolumn{4}{c}{\textbf{20m}} & & \multicolumn{4}{c}{\textbf{30m}} & & \multicolumn{4}{c}{\textbf{40m}} & & \multicolumn{4}{c}{\textbf{50m}} \\
\cmidrule{2-5} \cmidrule{7-10} \cmidrule{12-15} \cmidrule{17-20} \cmidrule{22-25}
         & Pr. & Rec. & F1 & MAE & & Pr. & Rec. & F1 & MAE & & Pr. & Rec. & F1 & MAE & & Pr. & Rec. & F1 & MAE & & Pr. & Rec. & F1 & MAE\\
\cmidrule[1.5pt]{2-5} \cmidrule[1.5pt]{7-10} \cmidrule[1.5pt]{12-15} \cmidrule[1.5pt]{17-20} \cmidrule[1.5pt]{22-25}
GM-only  & 82.3 & 69.0 & 75.1 & 0.54  & & 78.5 & 65.8 & 71.6 & 0.42 & & 75.5 & 46.8 & 57.8  & 0.76 &  & 55.6 & 29.8 & 38.9 & 0.99 & & 27.1 & 22.1 & 24.3 & 1.34 \\
LI-only  & 85.1 & 65.1 & 73.9 & 0.32 & & 83.1 & 64.2 & 72.0 & 0.35 & & 86.1 & 64.0 & 71.3 & 0.49 & & 74.5 & 57.8 & 65.1 & 0.50 & & 62.5 & 39.5 & 48.4 & 0.45\\
\bottomrule
\end{tabular}}
\caption{Ablation study on sensor measurements received by individual sensors. GM-only stands for only ground markings channel representing the processed information received from stereo camera and LI-only represents the lidar intensity channel.}
\label{table:singlesensor}
\end{table*}

In Table \ref{table:leave_one_out} we further perform a leave-one-out experiment, where we use all input channels except one. This is an indicator of the contribution of each channel to the overall results and whether there is redundant information across channels. It can be seen that removing lidar intensity results in the biggest performance drop across all distance ranges, implying it to be the most critical source of information. Ground markings from the camera seem to add overall value only in the near field, and occupancy information mostly contributes to the far distance cases where no markings can be observed explicitly anymore by camera or lidar. Similar to the ground markings provided through camera, traffic history mainly has positive influence in the near field. This is attributed to the fact that vehicles in the near range with respect to the ego-vehicle are tracked more robustly and hence provide richer information on the direction of traverse in the intersection area. This channel specifically represents implicit information on the existence of intersection area by outlining the trajectories perpendicular to the ego-vehicle\textquotesingle s direction representing cross traffic. 
\begin{table*}[htbp]
\centering
\resizebox{0.67\linewidth}{!}
{\begin{tabular}{lcccccccccccccc}
\toprule
Distance & \multicolumn{2}{c}{10m} &     & \multicolumn{2}{c}{20m}   &   & \multicolumn{2}{c}{30m}      & & \multicolumn{2}{c}{40m}  &    & \multicolumn{2}{c}{50m}      \\
\cmidrule{2-3} \cmidrule{5-6} \cmidrule{8-9} \cmidrule{11-12} \cmidrule{14-15}
         & F1    & MAE  & & F1   & MAE  & & F1            & MAE  &         & F1            & MAE      &     & F1            & MAE           \\
\cmidrule{2-3} \cmidrule{5-6} \cmidrule{8-9} \cmidrule{11-12} \cmidrule{14-15}
all-ELE  & \textbf{80.4}          & 0.32          & & 79.4          & 0.36          & & 75.2          & 0.42          & & 67.5          & 0.53          & & 55.0          & 0.57          \\
all-LI   & 81.7 & \textbf{0.50} & & 80.0          & \textbf{0.56} & & \textbf{66.8} & \textbf{0.74} & & \textbf{61.9} & \textbf{0.89} & & 48.3          & \textbf{0.91} \\ 
all-GM   & 81.9          & 0.32          & & \textbf{76.6} & 0.30          & & 71.7          & 0.45          & & 69.6          & 0.50          & & 58.7          & 0.50          \\ 
all-OCC  & 82.2          & 0.29          & & 81.4          & 0.28          & & 72.7          & 0.50          & & 67.2          & 0.44          & & \textbf{44.7} & 0.56          \\ 
all-SEM  & 84.9          & 0.30          & & 78.7          & 0.34          & & 75.3          & 0.49          & & 67.3          & 0.56          & & 59.6          & 0.53          \\
all-TH  & 81.8       & 0.34         & & 77.8          & 0.34          & & 75.6         & 0.37          & & 69.5          & 0.58         & & 55.4          & 0.54          \\
\bottomrule
\end{tabular}}
\caption{Ablation study of channels contributions, where all input channels \emph{except} the named one have been used. Abbreviations of the input channels are: ELE (Elevation map), LI (lidar intensity), GM (ground markings), OCC (occupancy), SEM (semantics ground), TH (traffic history).}
\label{table:leave_one_out}
\end{table*}
\subsection{Visual Interpretation}
Although over the past few years deep learning has been the key driving force in self-driving industry, it also raises safety and security issues due to the lack of interpretability and transparency \cite{ma2018secure}. Stop line detection is a safety-critical application, and hence understanding how the network retrieves information from different channels and makes decisions is very crucial. Therefore, to get a better understanding of network\textquotesingle s behavior, we employ an activation-based spatial attention mapping function \cite{pay_attetnion} that takes absolute value of a hidden neuron activation to indicate the importance of that neuron with respect to the specific input.

\begin{figure*}[htbp]
\centering
    \begin{subfigure}[b]{\linewidth}
         \centering
         \includegraphics[width=0.95\linewidth]{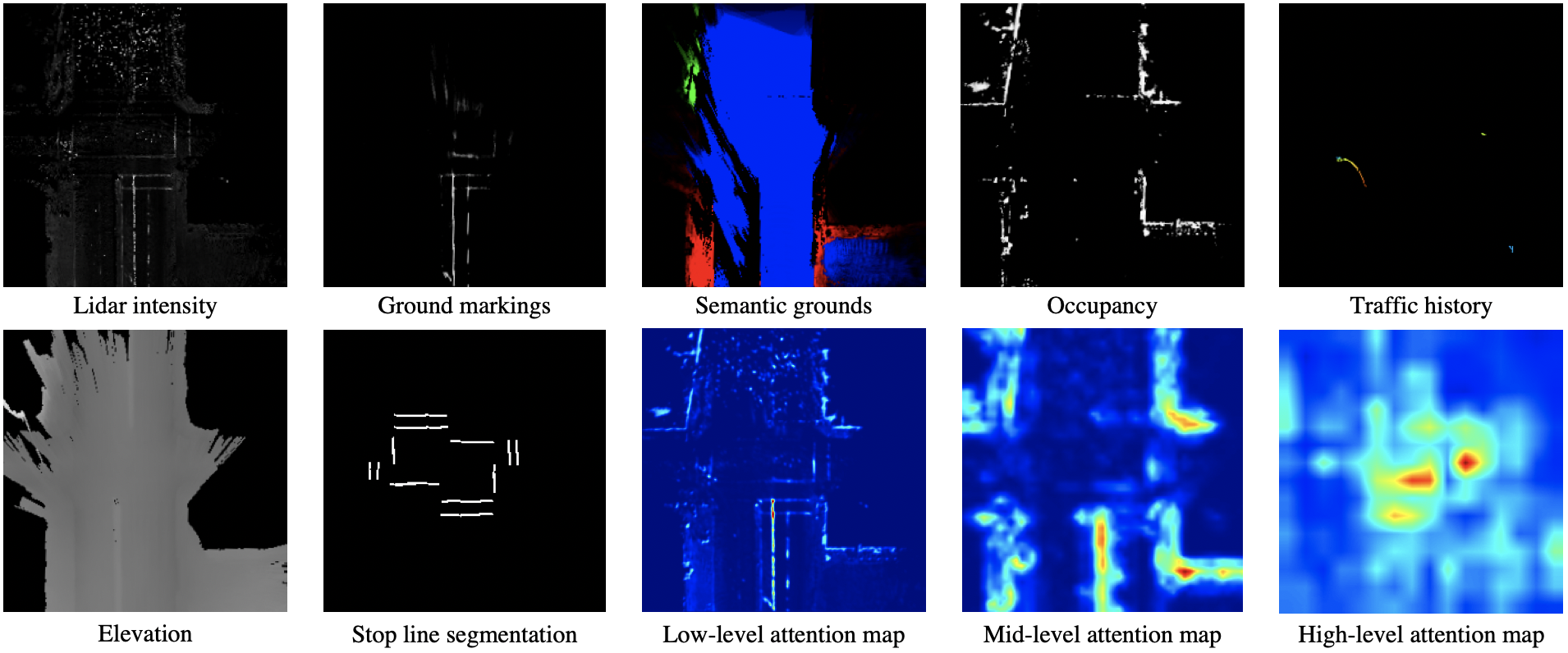}
         \caption{}
     \end{subfigure}
     \begin{subfigure}[b]{\linewidth}
         \centering
         \includegraphics[width=0.95\linewidth]{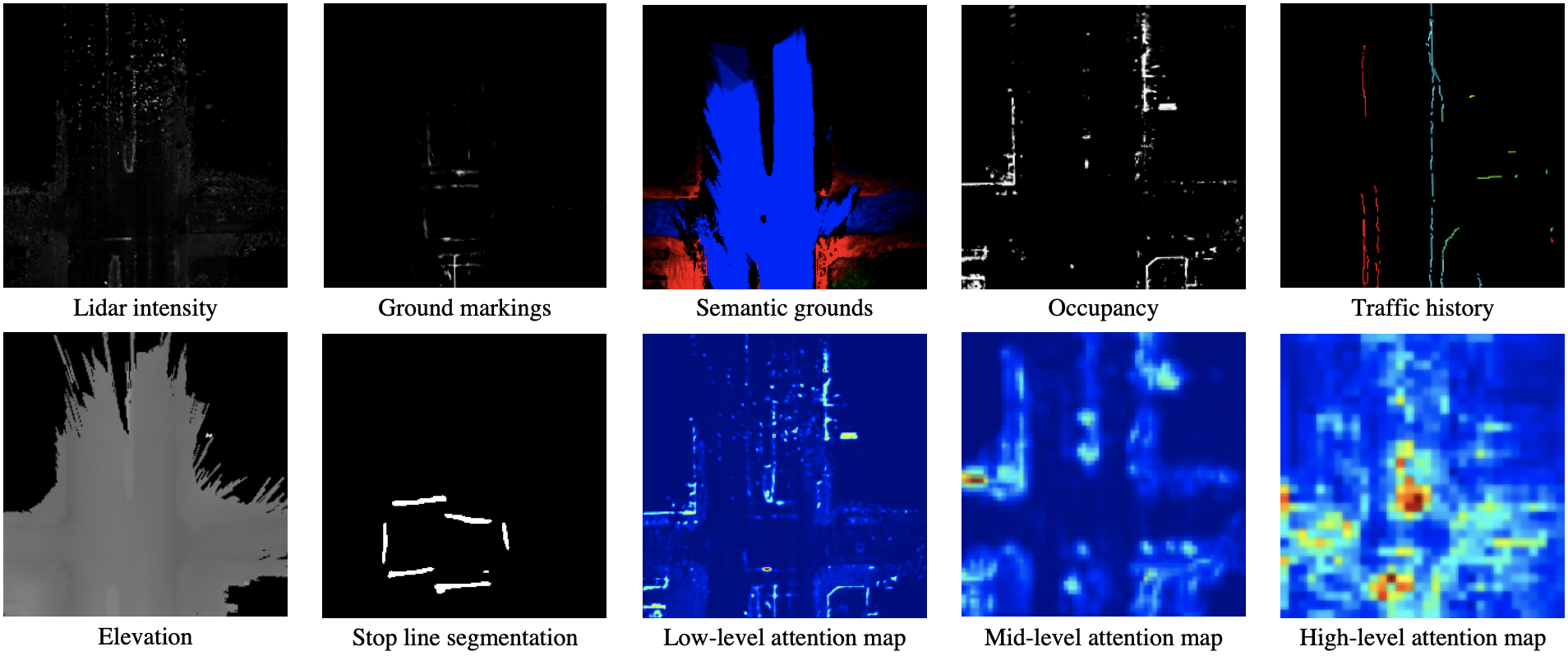}
         \caption{}
     \end{subfigure}
\caption{Intermediate visualizations of three different levels of the proposed network along with the corresponding input channels and inferred segmentations for two sample frames.}
\label{fig:UNet_visualize}
\end{figure*}

Figure \ref{fig:UNet_visualize} shows the input channels of an intersection scenario and the corresponding activation-based attention maps from low-level to high-level convolution layers. The attention map of the low-level intermediate layer is quite similar with a linear combination of different input channels, therefore we can conclude that the low-level convolution filters perform as a dynamic learnable weights assigner. It is quite clear to observe that lidar intensity, ground markings and occupancy are taking the majority weights and thus their qualities are relatively more important, which is consistent with our ablation study result in the previous section. The middle-level attention map is visually like a zoomed-in version of the low-level attention map. It solely focuses on the central part of the intersection and filters out the irrelevant area. Thus this layer functions as an intersection detector and passes the intersection area to next layer. Finally, the high-level convolution layer will extract abstract features and highlight the rough stop line area.

In summary, to detect the stop lines, our neural network first retrieves the most crucial information from multiple inputs to localize the intersection area and then infers the stop line positions from the intersection structure. This explains the reason that our approach can predict the stop lines even without observing the ground markings.
\section{Conclusion}
In this paper we proposed a holistic deep neural network-based approach to stop line detection that approaches the problem by combining explicit and implicit features of intersections, collected from multiple sensors. Our theoretical analysis and quantitative results show that using camera data alone, the detection range of stop lines is fairly limited and that adding additional features such as occupancy grid maps to implicitly detect the intersection structure helps to increase detection range.
The quantitative results also align well with the performed visual network introspection, which shows that the model effectively learns a weighted combination of our inputs on low-level feature channels and moves towards detecting intersections as a whole on higher-level feature channels.

Finding road intersections in advance is crucial for navigation and path planning of autonomous vehicles, especially when there is no position or geographic auxiliary information available. Therefore, having a system with capability of detecting stop line areas despite having occlusion or damaged road painting is very essential for the SDC to function properly. Such system would be also beneficial for the automatic creation of HD maps, as well as validation and change detection in the existing maps.

It should be noted that the position of stop signs and traffic lights are also strong additional cues to learn where to stop in an intersection. Since the focus of this work was on the importance of fused information received from different sensor measurements on the grid map level, it was not integrated as part of the system. In the future, however, we will employ such meta features to facilitate the learning process.
\bibliographystyle{IEEEtran}
\bibliography{IEEEfull}
\end{document}